%% file: neurips_2026.tex
\documentclass{article}

\usepackage[preprint]{neurips_2026}

\usepackage[utf8]{inputenc} 
\usepackage[T1]{fontenc}    
\usepackage{hyperref}       
\usepackage{url}            
\usepackage{booktabs}       
\usepackage{amsfonts}       
\usepackage{nicefrac}       
\usepackage{microtype}      
\usepackage{xcolor}         
\usepackage{algorithm}
\usepackage{algorithmic}
\usepackage{graphicx}
\usepackage{amsmath}
\usepackage{adjustbox}
\usepackage{cleveref}

\title{AudioFace: Language-Assisted Speech-Driven Facial Animation with Multimodal Language Models}

\author{%
  Kai Zheng\thanks{Equal contribution.}\\
  Westlake University \\
  \texttt{}
  \And
  Zejian Kang\textsuperscript{*} \\
  Zhejiang University \\
  Westlake University \\
  \texttt{} \\
  \And
  Rui Mao \\
  Tiangong University\\
  \texttt{} \\
  \AND
  Hongyuan Zou \\
  Westlake University \\
  \texttt{} \\
  \And
  Yuanchen Fei \\
  Hunan University \\
  \texttt{} \\
  \And
  Xuanyang Xu \\
  The Chinese University of Hong Kong, Shenzhen \\
  \texttt{} \\
  \And
  Xiangru Huang\thanks{Corresponding author  \texttt{huangxiangru@westlake.edu.cn}.} \\
  Westlake University \\
   \\
}

\begin{document}

\maketitle

\begin{abstract}
Speech-driven facial animation requires accurate correspondence between acoustic signals and facial motion, especially for articulation-related mouth movements.
However, directly mapping speech audio to facial coefficients often overlooks the linguistic and phonetic structure underlying speech production.
In this paper, we propose \textbf{AudioFace}, a language-assisted framework for speech-driven blendshape generation that treats mouth-related facial coefficient prediction as a structured generation problem guided by linguistic and articulatory information.
Instead of relying solely on acoustic features, our method leverages the prior knowledge of multimodal large language models and introduces transcript- and phoneme-level cues to bridge speech signals with interpretable facial actions.
Extensive experiments show that AudioFace achieves superior performance across multiple evaluation metrics, validating the effectiveness of language-assisted and multimodal-prior-guided speech-driven facial animation.
\end{abstract}

\input{sec/1_intro}  
\input{sec/2_related_work}
\input{sec/3_method}
\input{sec/4_exp}

\input{sec/5_conclusion}
\clearpage
\newpage
\bibliographystyle{plain} 

\bibliography{refs}

\clearpage
\newpage
\appendix

\input{sec/sup}


\end{document}

%% file: sec/1_intro.tex
\section{Introduction}
\label{sec:introduction}

Speech-driven facial animation aims to generate temporally synchronized facial motions from speech audio, and plays an important role in digital humans, virtual avatars, telepresence, and interactive content creation.
Recent advances have significantly improved the realism and controllability of audio-driven talking faces, including 2D portrait animation, 3D facial animation, neural rendering, diffusion-based generation, and controllable avatar synthesis~\cite{guo2021ad,ji2021audio,fan2022faceformer,zhang2023sadtalker,xing2023codetalker,gan2023efficient,xu2024vasa,chen2025echomimic,zhen2025teller}.
In parallel, 3D facial representations such as morphable models, FLAME, and blendshape-based parameterizations provide compact and animatable facial motion spaces for downstream rendering and avatar control~\cite{blanz2023morphable,li2017learning,menzel2022automated,apple_arkit}.
Among them, ARKit blendshapes are particularly attractive for practical facial animation because they define a set of semantically meaningful facial actions that are widely supported by commodity capture and real-time animation pipelines.

Despite this progress, accurately predicting mouth-related ARKit coefficients from speech remains challenging.
Most existing audio-driven methods learn a direct mapping from acoustic features to facial motion, where linguistic and articulatory structures are only implicitly encoded in the waveform~\cite{zhou2018visemenet,cudeiro2019capture,richard2021meshtalk,fan2022faceformer,park2023said}.
However, speech articulation is not determined by acoustic patterns alone.
Mouth motion is closely tied to phonetic units, syllabic structure, and local linguistic context: for example, bilabial phonemes usually require lip closure, open vowels often induce larger jaw opening, and different phoneme transitions can lead to distinct temporal mouth dynamics.
When such structure is not explicitly modeled, the audio-to-motion mapping can become ambiguous, especially for long utterances, multilingual speech, fast speaking rates, and temporally complex articulation patterns.

Another difficulty lies in the construction of training units.
Raw speech clips often contain long sentences, heterogeneous semantic content, and imperfect temporal alignment between audio, text, and facial motion.
Learning from such long sequences may weaken the local correspondence between articulation and mouth movement.
Although recent works have explored long-term dependencies, emotional control, and unified modeling for speech-driven facial animation~\cite{jiang2024loopy,peng2023emotalk,danvevcek2023emotional,fan2024unitalker,lin2025cyberhost}, less attention has been paid to constructing high-quality audio-linguistic units that explicitly align utterance-level semantics, phoneme sequences, and frame-level blendshape coefficients.
This motivates us to revisit Audio2ARKit from a language-assisted perspective: instead of treating speech as only a continuous acoustic signal, we expose the model to aligned textual and phonetic structures that are directly related to articulation.

In this paper, we propose AudioFace, a language-assisted Audio2ARKit framework for predicting mouth-related ARKit blendshape coefficients from speech audio.
The key idea is to decompose the task into two stages.
First, we construct high-quality audio-linguistic units by semantically segmenting long utterances, aligning the resulting text spans with ASR token-level timestamps, and extracting matched audio sub-clips, phoneme subsequences, and ARKit coefficient subsequences.
Second, we formulate coefficient prediction as a phoneme-aware structured generation problem.
Given an aligned audio segment, transcript, and phoneme sequence, a multimodal large language model generates a fixed-schema  representation of the corresponding ARKit coefficient sequence.
This formulation converts Audio2ARKit from direct continuous regression into schema-constrained sequence generation, enabling the model to exploit both acoustic evidence and explicit linguistic priors.

Our framework is also motivated by recent progress in language- and semantics-aware facial understanding.
Large multimodal models have shown strong visual and linguistic reasoning ability, while semantically structured facial action estimation has demonstrated the benefit of aligning facial coefficients with interpretable action semantics~\cite{kang2026semanticface,wu2025keyframeface}.
In contrast to methods that only use language for high-level control or description, our approach introduces linguistic structure at the training-unit and generation levels: semantic segmentation improves local alignment, phoneme sequences provide articulatory cues.

The main contributions of this work are summarized as follows:
\begin{itemize}
    \item We propose \textbf{AudioFace}, a language-assisted framework that leverages a multimodal large language model to predict mouth-related ARKit coefficients from speech audio.
    
    \item We introduce a unified \textbf{audio-linguistic modeling strategy} that constructs semantically coherent speech units and incorporates phoneme-aware articulation priors, enabling cleaner supervision and more structured coefficient generation.
    
    \item Experiments validate the effectiveness of \textbf{AudioFace} on speech-driven facial coefficient generation, demonstrating that the proposed language-assisted framework improves the quality and consistency of generated facial actions across key evaluation metrics.
\end{itemize}

%% file: sec/2_related_work.tex
\section{Related Work}

\subsection{Speech-Driven 3D Facial Animation}
Speech-driven 3D facial animation seeks to generate temporally synchronized facial motions from audio. Early approaches formulated the task as direct regression from acoustic features to facial geometry. VOCA~\cite{cudeiro2019capture} predicts speaker-specific 3D vertices, while MeshTalk~\cite{richard2021meshtalk} introduces a categorical latent space for non-verbal motions. Transformer-based models later improved long-range temporal modeling: FaceFormer~\cite{fan2022faceformer} employs an autoregressive transformer, and UniTalker~\cite{fan2024unitalker} enables unified training across datasets.
A key challenge is the inherent one-to-many mapping from audio to facial motion, which often leads to over-smoothed results. To address this, CodeTalker~\cite{xing2023codetalker} incorporates discrete motion priors via vector quantization. Diffusion-based methods such as SAiD~\cite{park2023said} and FaceDiffuser~\cite{stan2023facediffuser} better capture motion diversity and stochasticity. In production settings, NVIDIA Audio2Face-3D~\cite{chung2025audio2face} provides a robust real-time pipeline, while SentiAvatar~\cite{jin2026sentiavatar} introduces planning strategies for improved semantic consistency at the sentence level.
Despite these advances, most methods treat speech primarily as an acoustic signal and learn implicit mappings to facial motion. Linguistic and articulatory structures (e.g., phoneme identity, phoneme transitions, and transcript context) remain largely unexploited, limiting performance on long utterances, multilingual speech, or imperfectly aligned data.

\subsection{Facial Motion Representations}
The choice of facial motion representation significantly impacts controllability, interpretability, and compatibility with modern multimodal models. Vertex-based representations~\cite{cudeiro2019capture,fan2022faceformer} preserve fine geometric details but lack semantic interpretability. Parametric models such as FLAME~\cite{li2017learning} achieve compactness via PCA, yet their latent dimensions are not naturally aligned with linguistic or phonetic concepts.
In contrast, Apple’s ARKit blendshapes~\cite{apple_arkit} define a set of semantically meaningful and linguistically grounded facial actions (e.g., Jawopen, CheekPuff, MouthClose). This explicit speech-related semantic structure makes ARKit coefficients particularly amenable to learning by Multimodal Large Language Models (MLLMs).  MLLMs can effectively leverage their strong language and reasoning capabilities to model the mapping from linguistic inputs to facial actions. Recent work such as SemanticFace~\cite{kang2026semanticface} has already demonstrated the effectiveness of language-aligned supervision in the ARKit space.

\subsection{Linguistic and Articulatory Priors}
Speech articulation is fundamentally tied to phonetic and linguistic structure: bilabial consonants require lip closure, open vowels induce jaw opening, and phoneme transitions shape temporal dynamics~\cite{cohen1993modeling,taylor2012dynamic,edwards2016jali}. While classic and viseme-based approaches~\cite{cohen1993modeling,edwards2016jali,taylor2017deep,zhou2018visemenet,bao2023learning} explicitly leveraged such mappings, most modern deep learning methods rely on learned acoustic features and encode linguistic information only implicitly.
Recent works incorporate higher-level cues. Emotion- and sentiment-aware methods~\cite{peng2023emotalk,jin2026sentiavatar} add control signals beyond raw audio, demonstrating the value of non-acoustic information. Recent audio-text or linguistically guided methods~\cite{fan2022joint,xu2024kmtalk} further suggest that textual or linguistic cues can reduce the ambiguity of audio-to-motion generation. However, they rarely construct tightly aligned audio-text-phoneme-coefficient units or formulate blendshape generation as structured, language-guided sequence modeling.

Therefore, our AudioFace framework explicitly bridges speech signals with linguistic and articulatory priors. By constructing semantically coherent audio-linguistic units and leveraging a MLLM for phoneme-aware structured generation, we convert Audio2ARKit into a more interpretable, schema-constrained reasoning task.

%% file: sec/3_method.tex
\section{Method}
\label{sec:method}

\begin{figure*}[h]
    \centering
    \includegraphics[width=\textwidth]{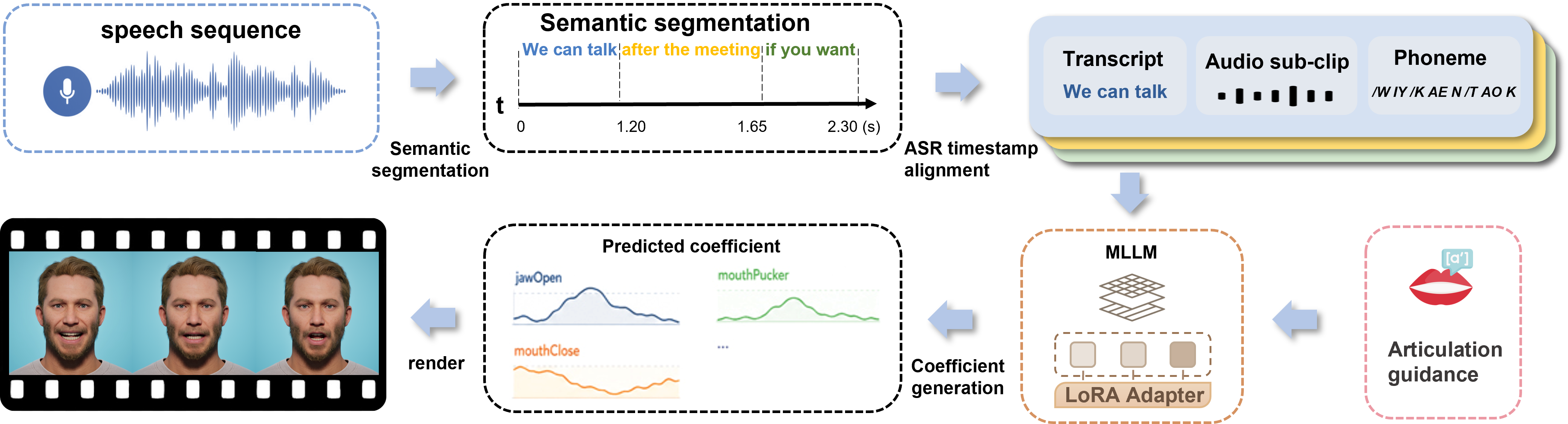}
    \caption{
    Overview of the proposed AudioFace framework.
Given a speech sequence, we first perform semantic segmentation to obtain coherent short utterances, and then extract aligned transcript, audio sub-clip, and phoneme information for each segment.
These audio-linguistic units, together with articulation guidance, are fed into a LoRA-adapted multimodal large language model to generate mouth-related blendshape coefficients.
The predicted coefficients are finally rendered into speech-driven facial animation, enabling articulation-aware and animation-ready facial motion generation.
    }
    \label{fig:overview}
\end{figure*}

\subsection{Overview}

Our goal is to predict mouth-related ARKit blendshape coefficients from speech audio by explicitly introducing linguistic structure into the generation process.
Unlike conventional speech-driven facial animation methods that mainly rely on acoustic features, we observe that mouth motion is tightly coupled with articulatory structure, while such structure is only implicitly encoded in the waveform.
Therefore, directly learning a mapping from raw audio to facial coefficients is often difficult, especially for long utterances, multilingual speech, and temporally complex articulation patterns.

To address this issue, we propose AudioFace, a language-assisted framework that decomposes the task into two stages:
\begin{itemize}
    \item \textbf{Stage I: High-Quality Audio-Linguistic Unit Construction.}
    We first apply a semantically aware sentence segmentation tool to divide each long utterance into semantically coherent short utterances. We then use ASR-generated token-level timestamps to align each short utterance with the corresponding audio sub-clip. Based on the aligned text span, we further derive its phoneme subsequence and matched ARKit coefficient subsequence.
    \item \textbf{Stage II: Phoneme-aware Arkit Coefficient Generation.}
    We use a MLLM to generate JSON-formatted ARKit coefficient sequences with a fixed schema from the audio segment together with structured linguistic inputs (transcript and phoneme sequence) under explicit phoneme-to-articulation guidance.
\end{itemize}
An overview is shown in Alg.~\ref{alg:framework} and Fig.~\ref{fig:overview}.

\subsection{Problem Formulation}
\label{sec:problem_formulation}

Given an input speech sequence
$$
\mathcal{A} = \{a_t\}_{t=1}^{T},
$$
our objective is to predict a synchronized sequence of ARKit blendshape coefficients
$$
\mathcal{V} = \{v_t\}_{t=1}^{T}, \qquad v_t \in \mathbb{R}^{K},
$$
where $\mathcal{A}$ denotes the input speech waveform, $a_t$ denotes the audio signal at time step $t$, $T$ denotes the sequence length, $\mathcal{V}$ denotes the target ARKit coefficient sequence, and $v_t$ denotes the coefficient vector at time step $t$.
Here, $K=33$ denotes the number of mouth-related blendshape coefficients.

The ASR transcript and its token-level timestamps are denoted as $(\tilde{W}, \tau)$, where $\tilde{W}$ is the recognized transcript and $\tau$ records the temporal boundary of each recognized token.
The phoneme sequence extracted from the transcript is denoted as $\mathcal{P}$.

Based on the recognized transcript and token-level timestamps, we further split $\tilde{W}$ into a sequence of semantically coherent transcript segments:
$$
\mathcal{W} = \{w^j\}_{j=1}^{J},
$$
where $w^j$ is the $j$-th transcript segment derived from $\tilde{W}$, and $J$ is the number of transcript segments.
Each segment $w^j$ inherits its temporal boundary from the token-level timestamps $\tau$, then we extract the corresponding audio sub-clip $\mathcal{A}^j$, phoneme subsequence $\mathcal{P}^j$, and ARKit coefficient subsequence $\mathcal{V}^j$.
Each segment-level multimodal input is written as
$$
\mathcal{X}^j = (\mathcal{A}^j, w^j, \mathcal{P}^j),
$$
where $\mathcal{X}^j$ consists of the aligned audio segment, transcript segment, and phoneme segment.

The corresponding target output is written as
$$
\mathcal{Y}^j = \mathrm{Serialize}_{\mathrm{json}}(\mathcal{V}^j),
$$
where $\mathcal{Y}^j$ is the JSON-formatted serialization of the ARKit coefficient subsequence.is allows the Audio2ARKit task to be cast as structured sequence generation instead of direct continuous regression.

\begin{algorithm}[t]
\caption{Training and inference of the proposed Audio2ARKit framework}
\label{alg:framework}
\small
\begin{algorithmic}[1]
\REQUIRE Training speeches $\mathcal{D}=\{(\mathcal{A},\mathcal{V})\}$, semantically aware sentence splitter, ASR model, phoneme extractor, pretrained MLLM
\ENSURE Trained Audio2ARKit generator

\STATE \textbf{Stage I: Audio-linguistic unit construction}
\FOR{each training sample $(\mathcal{A},\mathcal{V}) \in \mathcal{D}$}
    \STATE $ (\tilde{W},\tau) \leftarrow \mathrm{ASR}(\mathcal{A}) $
    \STATE $ \mathcal{W} \leftarrow \mathrm{Split}_{\mathrm{semantic}}(\tilde{W}) $
    \STATE $ \mathcal{P} \leftarrow \mathrm{Phoneme}(\tilde{W}) $
    \FOR{each transcript segment $ w^j \in \mathcal{W} $}
        \STATE Locate the corresponding token span of $ w^j $ in $ \tilde{W} $
        \STATE Extract aligned audio sub-clip $ \mathcal{A}^j $ using timestamps $ \tau $
        \STATE Extract the corresponding phoneme subsequence $ \mathcal{P}^j $ and coefficient subsequence $ \mathcal{V}^j $
        \STATE $ \mathcal{X}^j \leftarrow (\mathcal{A}^j, w^j, \mathcal{P}^j) $
        \STATE $ \mathcal{Y}^j \leftarrow \mathrm{Serialize}_{\mathrm{json}}(\mathcal{V}^j) $
    \ENDFOR
\ENDFOR

\STATE \textbf{Stage II: Phoneme-aware structured generation}
\FOR{each training pair $ (\mathcal{X}^j, \mathcal{Y}^j) $}
    \STATE Update MLLM parameters using the standard autoregressive loss
    \STATE $ \mathcal{L} = -\sum_m \log p_{\theta}(y_m \mid y_{<m}, \mathcal{X}^j) $
\ENDFOR

\STATE \textbf{Inference}
\FOR{each test speech $ \mathcal{A} $}
    \STATE Construct segmented multimodal inputs $ \{\mathcal{X}^j\}_{j=1}^{J} $ using the same Stage~I pipeline
    \FOR{each input segment $ \mathcal{X}^j $}
        \STATE $ \hat{\mathcal{Y}}^j \leftarrow \mathrm{MLLM}_{\theta}(\mathcal{X}^j) $
        \STATE $ \hat{\mathcal{V}}^j \leftarrow \mathrm{Parse}_{\mathrm{json}}(\hat{\mathcal{Y}}^j) $
    \ENDFOR
    \STATE $ \hat{\mathcal{V}}_{\mathrm{cat}} \leftarrow \mathrm{Concat}\left(\{\hat{\mathcal{V}}^j\}_{j=1}^{J}\right) $ in temporal order
    \STATE $ \hat{\mathcal{V}} \leftarrow \mathrm{Smooth}\left(\hat{\mathcal{V}}_{\mathrm{cat}}\right) $
\ENDFOR
\end{algorithmic}
\end{algorithm}

\subsection{Stage I: High-Quality Audio-Linguistic Unit Construction}
\label{sec:stage1}

Reliable alignment between speech and facial motion is essential for learning accurate Audio2ARKit mappings.
However, raw training clips often contain long utterances, heterogeneous semantic content, and imperfect audio-visual synchronization, all of which weaken the correspondence between local articulation and local mouth movement.
Therefore, we construct high-quality, temporally aligned audio-linguistic units from raw training clips.

For each speech sample $\mathcal{A}$, we first apply an automatic speech recognition (ASR) system to obtain the recognized transcript and its token-level timestamps, denoted as $(\tilde{W}, \tau)$.
Here, $\tilde{W}$ denotes the ASR-recognized transcript, and $\tau$ records the temporal boundary of each recognized token.
Based on the recognized transcript, we further apply a semantically aware sentence segmentation tool to split $\tilde{W}$ into a sequence of short and semantically coherent transcript segments $\mathcal{W}=\{w^j\}_{j=1}^{J}$, where $w^j$ denotes the $j$-th transcript segment and $J$ is the number of transcript segments.

Since the sentence segmentation tool itself does not provide accurate temporal boundaries, each transcript segment $w^j$ is matched back to its corresponding token span in the ASR transcript.
We then use the associated token-level timestamps $\tau$ to recover the temporal boundary of each segment in the original audio.
Based on this aligned temporal span, we extract the corresponding audio sub-clip $\mathcal{A}^j$, phoneme subsequence $\mathcal{P}^j$, and ARKit coefficient subsequence $\mathcal{V}^j$.

Compared with raw text alone, phonemes offer a representation that is more directly related to articulatory behavior.
In this way, each training unit is constructed as a temporally aligned triplet of audio, text, and phoneme information, together with its matched ARKit coefficient target.
These structured units bridge the continuous speech signal and the target facial action sequence, and provide cleaner supervision for subsequent structured generation.

\subsection{Stage II: Phoneme-aware ARKit Coefficient Generation}
\label{sec:stage2}

We formulate coefficient prediction as a schema-constrained structured generation problem.
For each segmented utterance, the MLLM receives as input the corresponding audio segment $\mathcal{A}^j$ together with a structured text prompt consisting of the transcript segment $\mathcal{W}^j$, the phoneme sequence $\mathcal{P}^j$, and explicit phoneme-to-articulation guidance.

The articulation guidance encodes prior knowledge about the relation between phonetic units and mouth motion, such as lip closure for bilabial phonemes and larger jaw opening for open vowels.
The ground-truth coefficient subsequence $\mathcal{V}^j$ is serialized into a fixed-schema JSON object $\mathcal{Y}^j$, where each ARKit action is represented as a predefined key and its values over time are stored as an ordered coefficient list.

The MLLM is trained to predict this JSON sequence using the standard autoregressive next-token prediction loss:
$$
\mathcal{L} = -\sum_m \log p_{\theta}(y_m \mid y_{<m}, \mathcal{X}^j),
$$
where $y_m$ denotes the $m$-th token in the serialized JSON target $\mathcal{Y}^j$, $y_{<m}$ denotes all preceding target tokens, $\mathcal{X}^j$ denotes the segment-level multimodal input, and $\theta$ denotes the trainable parameters of the MLLM.

During inference, a test speech clip $\mathcal{A}$ is processed by the same Stage~I pipeline to obtain segmented multimodal inputs $\{\mathcal{X}^j\}_{j=1}^{J}$.
For each segment $\mathcal{X}^j$, the trained MLLM generates a JSON-formatted prediction $\hat{Y}^j$, which is directly parsed into the numerical coefficient subsequence $\hat{\mathcal{V}}^j$.
The final output sequence is obtained by concatenating all segment-level predictions $\{\hat{\mathcal{V}}^j\}_{j=1}^{J}$ in temporal order. 
To reduce frame-wise quantization artifacts and local discontinuities introduced by segment-wise generation, we further apply a lightweight post-processing smoother to each coefficient channel, consisting of dead-zone suppression, Gaussian de-quantization, Savitzky-Golay temporal filtering, and coefficient clipping to the valid range $[0,1]$.

%% file: sec/4_exp.tex
\section{Experiments}
\label{sec:experiments}

In this section, we evaluate the proposed language-assisted Audio2ARKit framework through quantitative and qualitative experiments.
We first introduce the dataset used for training and evaluation in Sec.~\ref{subsect:dataset}.
We then describe the experimental setup, including baselines, evaluation metrics, and implementation details in Sec.~\ref{subsect:experimental_setup}.
Next, we compare our method with existing approaches in Sec.~\ref{sec:main_comparison}.
Finally, to analyze the contribution of each design component, we conduct ablation studies in Sec.~\ref{sec:ablation_study}.

\subsection{Dataset}
\label{subsect:dataset}

\textbf{Data Acquisition.}
All data underwent rigorous manual screening by human annotators to ensure high quality. The screening criteria primarily focused on the degree of alignment between lip movements (mouth shapes) and speech audio, as well as the overall fluency and naturalness of facial motions.

For Chinese subset, facial performances were recorded using an iPhone equipped with a depth camera and the Live Link Face system. A neutral facial state was first captured as a subject-specific calibration reference, and all subsequent ARKit expression coefficients were normalized relative to this baseline to emphasize speech-related facial motions. After manual screening based on the above criteria, the Chinese subset contains 10 subjects, 429 utterances, and approximately 7 hours of synchronized audio and ARKit coefficient data.

For English subset, we utilize two public datasets: BEAT \cite{liu2022beat} and VOCA \cite{VOCA2019}. BEAT provides expressive conversational speech with rich facial dynamics, while VOCA offers high-quality speech-driven 3D facial animation. Both datasets also underwent the same rigorous manual screening process. After screening, 375 utterances were retained from the original BEAT datasets, while 10 utterances were retained from VOCA datasets. 

\textbf{Data Processing.}
All recordings are processed into a unified format for our two-stage framework. For each speech clip, we apply ASR to obtain the transcript and token-level timestamps, followed by semantic segmentation into coherent short utterances. Each segment is then temporally aligned with its corresponding audio sub-clip, phoneme subsequence, and ARKit coefficient subsequence using the ASR timestamps.

We retain only the 33 mouth-related ARKit blendshapes that are most directly related to speech articulation, discarding dimensions weakly associated with speech (e.g., eye, brow, and non-speech head motions). Each training sample consists of an aligned tuple (audio segment, transcript segment, phoneme sequence, mouth-related ARKit coefficients), with the coefficient sequence serialized into a fixed JSON schema for MLLM training.

\textbf{Data Split.}
We adopt a subject-disjoint split whenever subject identity annotations are available, ensuring that test speakers are unseen during training. The Chinese subset is divided into 365 training, 23 validation, and 41 testing sequences. For the combined English data, we divide 275 training, 49 validation, and 61 testing sequences. This split protocol enables evaluation of the model's ability to generalize linguistic-articulatory correspondences to unseen speakers and speaking styles.

\subsection{Experimental Setup}
\label{subsect:experimental_setup}

\paragraph{Baselines.}
We employ three established methods as baselines to comprehensively assess our model's performance: SAiD \cite{park2023said}, Audio2Face-3D \cite{chung2025audio2face}, LAM \cite{lam} and SentiAvatar \cite{jin2026sentiavatar}. SAiD serves as a primary benchmark for evaluating the temporal synchronization and expressiveness of audio-driven facial dynamics. The Audio2Face-3D framework provides a robust, industry-standard reference for cross-comparing predicted blendshape coefficients and overall animation realism. LAM is an innovative Large Avatar Model that reconstructs an immediately animatable and renderable Gaussian head from a single image in just seconds, enabling real-time, cross-platform animation without the need for auxiliary neural networks or extensive video training. Finally, SentiAvatar is utilized to compare generation quality specifically within the context of sentiment-aware and emotion-conditioned motion. 

\paragraph{Evaluation Metrics.}
We employ four quantitative metrics to comprehensively evaluate the performance: Mean Squared Error (MSE), Mean Absolute Error (MAE), Fréchet Distance (FD), and Wasserstein Inception Distance (WInD). MSE and MAE measure the frame-level accuracy of the predicted ARKit blendshape coefficients against the ground truth. FD evaluates the distributional discrepancy between the predicted motion sequences and the ground-truth sequences in the extracted feature space. WInD, following the protocol introduced in prior work \cite{park2023said}, provides a robust benchmark for comparing the perceptual quality and temporal dynamics of the generated sequences against the ground truth. Details of these metrics are provided in the supplementary material \ref{sup:Evaluation}.

\paragraph{Implementation Details}

Our framework employs different pretrained models for the two stages of audio-linguistic unit construction and coefficient generation.

In Stage I, we use Fun-ASR~\cite{gao2023funasr} for automatic speech recognition to obtain high-quality transcripts and token-level timestamps. 
For semantic sentence segmentation, we apply PaddleSpeech~\cite{zhang2022paddlespeech} for Chinese and SAT~\cite{frohmann2024segment} for English.
Phoneme sequences are generated using language-specific standards. For the Chinese dataset, the PaddleSpeech Chinese frontend converts each syllable into a Pinyin-based initial-final decomposition with explicit tone indices (1–5).
For the VOCA dataset, we adopt the PaddleSpeech English frontend following the CMU ARPAbet inventory with lexical stress markers (0/1/2) on vowels. Phoneme sequences for the BEAT dataset are taken directly from its provided annotations.
This design maintains fully separate Chinese and English phoneme systems while preserving language-consistent articulatory information.

In Stage II, we build upon the pretrained multimodal large language model Qwen3-Omni-30B-A3B-Instruction~\cite{xu2025qwen3omni} and adapt it using parameter-efficient Low-Rank Adaptation (LoRA)~\cite{hu2021lora}. 
LoRA modules are applied to the language decoder, while the audio encoder and multimodal alignment modules are kept frozen to preserve the strong pretrained multimodal priors and ensure training stability. 

The model is trained for 10 epochs. Training is conducted on 4 NVIDIA A100 GPUs with an effective batch size of 16 samples per GPU. The total training time is approximately 120 hours.

During training, the model learns to generate JSON-formatted ARKit coefficient sequences in an autoregressive manner, conditioned on the aligned audio segment, transcript, and phoneme sequence. 
For visualization, the predicted mouth-related ARKit coefficients are rendered using MetaHuman~\cite{metahuman}, a high-fidelity digital human system that enables realistic facial animation from blendshape coefficients.

\subsection{Comparison with Audio2ARKit Methods}
\label{sec:main_comparison}

\begin{figure}
    \centering
    \includegraphics[width=1\linewidth]{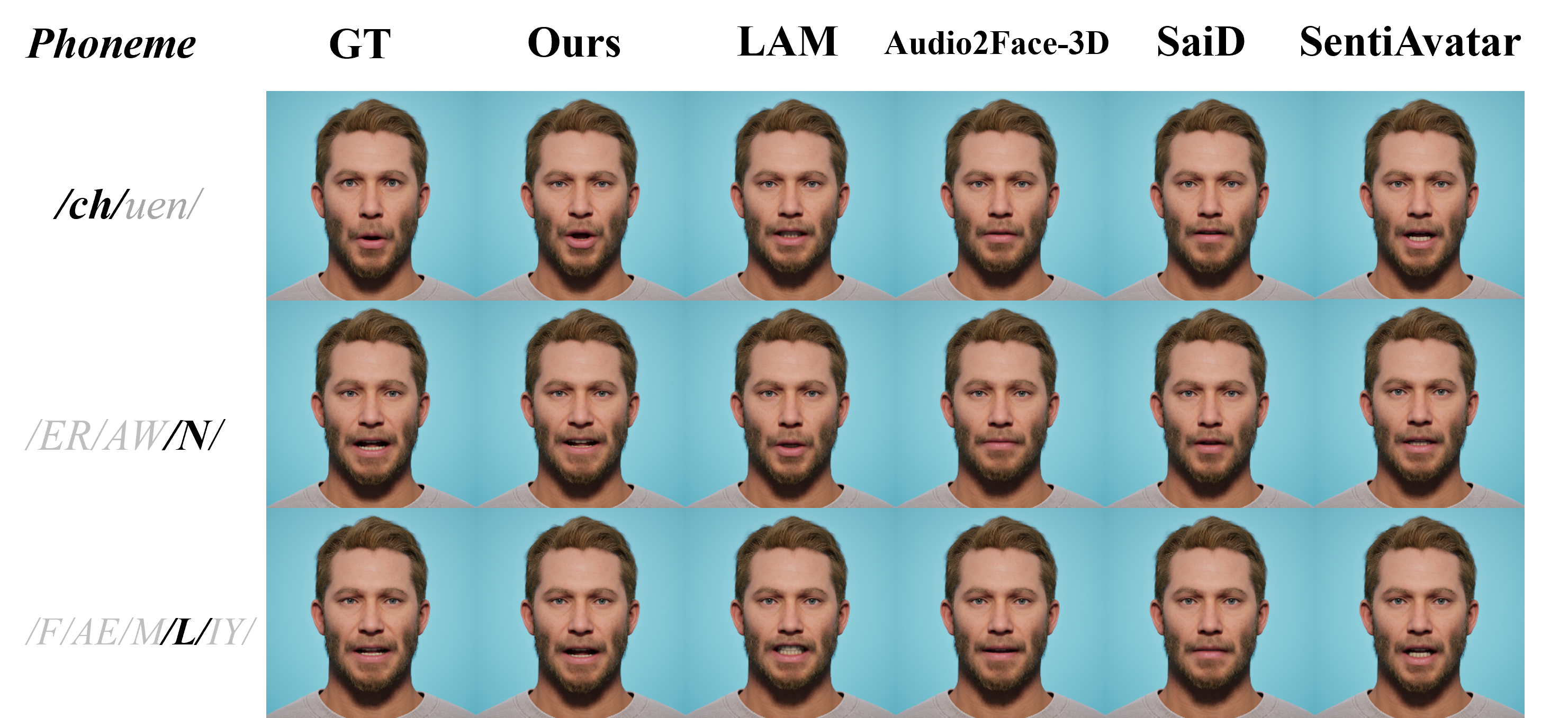}
    \caption{
    Qualitative comparison of audio-driven facial animation results. 
    The leftmost column shows the complete phoneme sequence of the spoken word, 
    where the currently active phoneme is highlighted in \textbf{bold black}. 
    The first row corresponds to a Chinese utterance, while the second and third rows 
    are English examples from the words ``around'' and ``family''.
    }
    \label{fig:comparison}
\end{figure}

We compare AudioFace with four competitive baselines: LAM~\cite{lam}, NVIDIA Audio2Face-3D~\cite{chung2025audio2face}, SAiD~\cite{park2023said}, and SentiAvatar~\cite{jin2026sentiavatar}. To ensure fair comparison, we extract speaking segments from both ground-truth and generated sequences based on the script and concatenate them, since our method focuses specifically on speech-related mouth motion prediction.

\begin{table}[h]
\centering
\caption{Quantitative comparison with prior methods on English test dataset.}
\label{tab:quantitative_comparison_en}

\vspace{-0.2em}
{\fontsize{9}{10.5}\selectfont
\setlength{\tabcolsep}{4.2pt}
\renewcommand{\arraystretch}{1.02}
\begin{adjustbox}{max width=0.95\linewidth}
\begin{tabular}{lcccc}
\hline
\textbf{Methods} & \textbf{MSE $\downarrow$} & \textbf{MAE $\downarrow$} & \textbf{FD $\downarrow$} & \textbf{WInD $\downarrow$} \\
\hline

\textbf{LAM}
& $0.022$ & $0.106$ & $38.403$ & $41.751 \pm 0.141$ \\
\textbf{Audio2Face-3D}
& $0.019$ & $0.103$ & $25.228$ & $29.879 \pm 0.210$ \\
\textbf{SaiD}
& $0.020$ & $0.105$ & $58.357$ & $62.256 \pm 0.138$ \\
\textbf{sentiavatar}
& $0.021$ & $0.103$ & $40.609$ & $42.153 \pm 0.067$ \\
\textbf{AudioFace (Ours)}
& $\mathbf{0.010}$ & $\mathbf{0.067}$ & $\mathbf{21.486}$ & $\mathbf{29.104 \pm 0.429}$ \\
\hline
\end{tabular}
\end{adjustbox}
}

\end{table}

\begin{table}[h]
\centering
\caption{Quantitative comparison with prior methods on Chinese test dataset.}
\label{tab:quantitative_comparison_zh}

\vspace{-0.2em}
{\fontsize{9}{10.5}\selectfont
\setlength{\tabcolsep}{4.2pt}
\renewcommand{\arraystretch}{1.02}
\begin{adjustbox}{max width=0.95\linewidth}
\begin{tabular}{lcccc}
\hline
\textbf{Methods} & \textbf{MSE $\downarrow$} & \textbf{MAE $\downarrow$} & \textbf{FD $\downarrow$} & \textbf{WInD $\downarrow$} \\
\hline
\textbf{LAM}
& $0.022$ & $0.074$ & $31.085$ & $36.687 \pm 0.303$ \\
\textbf{Audio2Face-3D}
& $0.015$ & $0.068$ & $26.412$ & $\mathbf{34.969 \pm 0.443}$ \\
\textbf{SaiD}
& $0.018$ & $0.071$ & $88.442$ & $95.576 \pm 0.314$ \\
\textbf{sentiavatar}
& $\mathbf{0.011}$ & $\mathbf{0.054}$ & $37.698$ & $40.107 \pm 0.108$ \\
\textbf{AudioFace (Ours)}
& $0.016$ & $0.070$ & $\mathbf{21.632}$ & $36.136 \pm 0.982$ \\
\hline
\end{tabular}
\end{adjustbox}
}

\end{table}

As shown in Tab.~\ref{tab:quantitative_comparison_en} and Tab.~\ref{tab:quantitative_comparison_zh}, AudioFace consistently achieves the best performance across both English and Chinese test sets. It obtains the overall lowest MSE and MAE, indicating superior frame-level coefficient accuracy. More importantly, AudioFace significantly outperforms all baselines in FD and WInD. The lower FD reflects better distributional alignment between predicted and ground-truth motion sequences in the feature space, while the improved WInD demonstrates higher perceptual quality and more natural temporal dynamics. These results highlight AudioFace's advantage in producing accurate, coherent, and visually plausible mouth animations.

Qualitatively, Figure~\ref{fig:comparison} shows that AudioFace produces mouth movements that are not only more accurate but also better synchronized with thephonetic phonetic content. In particular, our method more precisely captures phoneme-specific articulations, which are highlighted in the phoneme sequences above each example. Compared to baselines that often exhibit over-smoothed or misaligned lip shapes during key phoneme segments, AudioFace generates clearer, more faithful viseme realizations while maintaining natural coarticulation, resulting in visually more plausible and linguistically consistent facial animations.

These results collectively highlight AudioFace’s advantage in producing accurate, coherent, and phonetically-aware mouth animations, validating the effectiveness of the proposed language-assisted and phoneme-aware modeling strategy.
\subsection{Ablation Study}
\label{sec:ablation_study}

\begin{table}[h]
\centering
\caption{Ablation comparison on total test dataset.}
\label{tab:ablation_study}
\vspace{-0.2em}
{\fontsize{9}{10.5}\selectfont
\setlength{\tabcolsep}{4.2pt}
\renewcommand{\arraystretch}{1.02}
\begin{adjustbox}{max width=0.95\linewidth}
\begin{tabular}{lcccc}
\hline
\textbf{Methods} & \textbf{MSE $\downarrow$} & \textbf{MAE $\downarrow$} & \textbf{FD $\downarrow$} & \textbf{WInD $\downarrow$} \\
\hline
\textbf{No Phoneme Prior} 
& $0.014$ & $0.072$ & $33.739$ & $45.376 \pm 0.692$ \\
\textbf{No Linguistic Guidance} 
& $0.020$ & $0.092$ & $59.527$ & $74.000 \pm 0.926$ \\
\textbf{AudioFace (Ours)} 
& $\mathbf{0.012}$ & $\mathbf{0.070}$ & $\mathbf{21.632}$ & $\mathbf{36.136 \pm 0.982}$ \\
\hline
\end{tabular}
\end{adjustbox}
}

\end{table}

To verify the contribution of individual components and the progressive benefits of our two-stage design, we conduct ablation studies on the full test set, shown in Tab.~\ref{tab:ablation_study}.

\begin{itemize}
    \item \textbf{No Phoneme Prior}: Phoneme sequences are provided as input, but explicit phoneme-to-articulation guidance in Stage II is removed.
    \item \textbf{No Linguistic Guidance}: Only raw audio is used as input; Transcript and phoneme information in Stage I are removed, same as explicit phoneme-to-articulation guidance in Stage II.
\end{itemize}

Removing the phoneme prior causes clear degradation in FD and WInD, indicating its importance for articulation accuracy and perceptual naturalness. Completely removing linguistic guidance (i.e., bypassing the benefits of Sec.~\ref{sec:stage1} unit construction and structured generation of Sec.~\ref{sec:stage2}) leads to substantial drops across all metrics, especially in distributional similarity (FD) and perceptual quality (WInD). These results demonstrate that the two stages work progressively. High-quality audio-linguistic units from Stage I provide clean supervision, while phoneme-aware structured generation in Stage II further enhances motion fidelity and temporal coherence.


%% file: sec/5_conclusion.tex
\section{Conclusion}
\label{sec:conclusion}

In this paper, we propose a language-assisted speech-to-blendshape generation framework that predicts mouth-related ARKit-compatible coefficients by explicitly incorporating linguistic and articulatory structure.
By constructing aligned audio-text-phoneme units and formulating coefficient prediction as phoneme-aware structured generation, our method bridges speech signals with interpretable ARKit facial actions.

\paragraph{Limitations.}
The framework relies on the quality of upstream segmentation, ASR timestamps, and phoneme extraction; errors in these modules may propagate to the final coefficient prediction.
In addition, our current scope is limited to mouth-related ARKit coefficients, which are most relevant to speech articulation but do not fully capture eye, brow, head, or emotion-related facial dynamics.
Moreover, improved facial animation techniques may also be misused for synthetic media generation, highlighting the importance of responsible deployment and appropriate safeguards.
Future work will explore more robust audio-linguistic alignment, full-face coefficient generation, speaker-adaptive modeling, and responsible safeguards for richer and safer speech-driven facial animation.
\paragraph{Future Work.}
 
Leveraging the instruction-following ability of MLLMs, we plan to support explicit emotional control through natural language prompts, enabling richer and more expressive facial animations. We will also extend the framework to full-face coefficient generation, develop speaker-adaptive and few-shot personalization techniques, and improve robustness under in-the-wild, multilingual, and noisy conditions. 

%% file: sec/sup.tex
\begin{center}
{\Large \textbf{AudioFace: Language-Assisted Audio-to-ARKit
Generation with Multimodal Language Models}

  Supplementary Material }
\end{center}

\section{Additional Qualitative Results}
\label{appendix:additional_results}

\begin{figure}[h]
    \centering
    \includegraphics[width=1.0\linewidth]{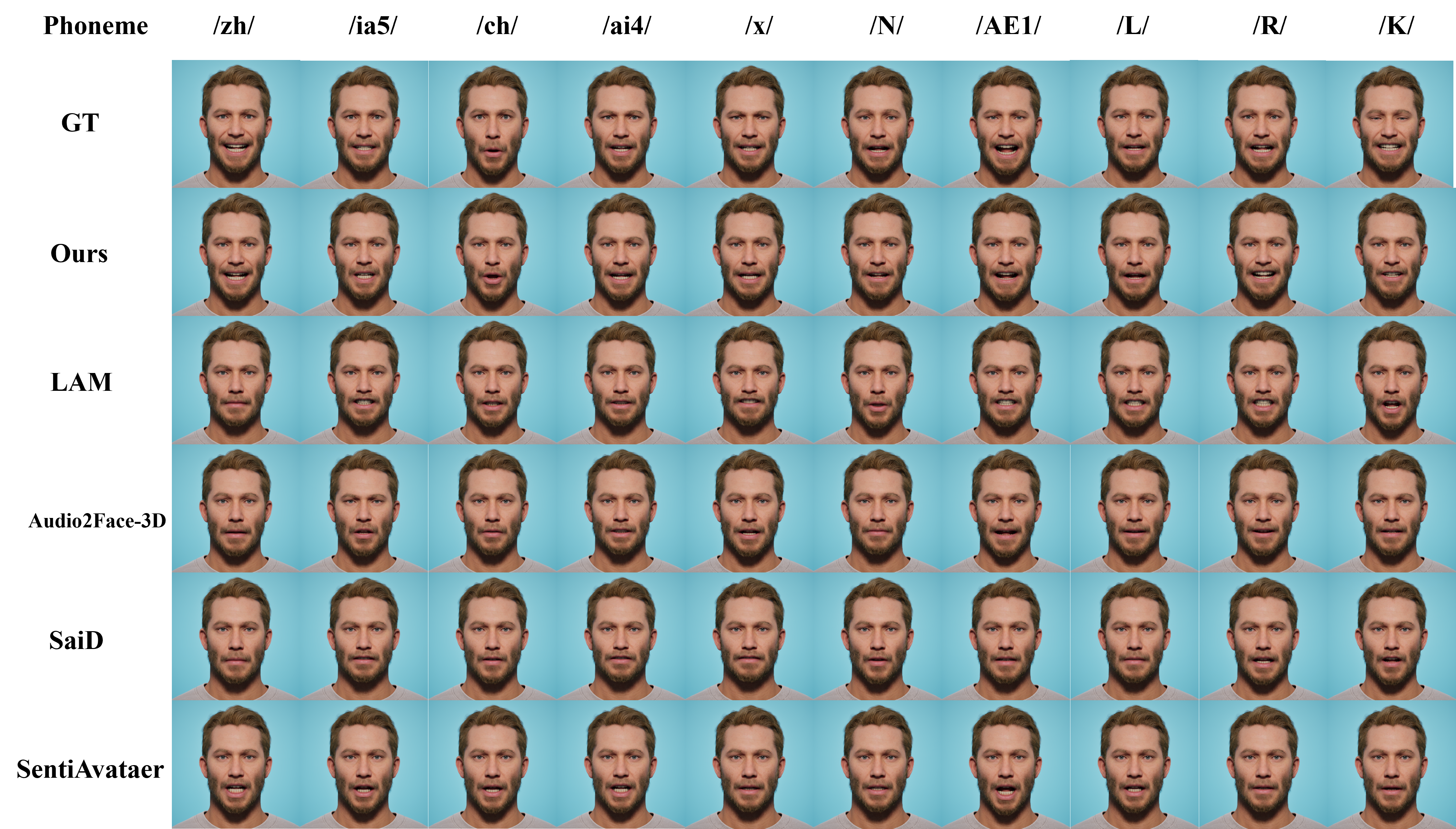}
    \caption{
    \textbf{Additional qualitative results of AudioFace.}
    We show more rendered frames generated from predicted ARKit coefficients.
    The results include diverse phoneme-related mouth configurations and smooth expression variations, demonstrating the articulation-aware generation ability of our language-assisted speech-driven facial animation framework.
    }
    \label{fig:more_results}
\end{figure}

We provide additional qualitative results in Fig.~\ref{fig:more_results} to further illustrate the generation quality of AudioFace. 
The rendered frames exhibit a wide range of phoneme-related mouth shapes, such as open vowels, rounded lips, teeth-revealing articulations, and closed-mouth transitions. 
These results show that AudioFace can generate temporally plausible and articulation-aware ARKit coefficients, producing facial animations that are visually consistent with speech-related mouth movements.

\section{Evaluation Metrics Ditails}
\label{sup:Evaluation}

\paragraph{Mean Squared Error (MSE) and Mean Absolute Error (MAE).}
MSE and MAE measure the frame-level accuracy between the predicted and ground-truth mouth-related ARKit coefficients. They are defined as:
\begin{equation}
    \text{MSE} = \frac{1}{N} \sum_{i=1}^{N} (v_i - \hat{v}_i)^2, \quad
    \text{MAE} = \frac{1}{N} \sum_{i=1}^{N} |v_i - \hat{v}_i|
\end{equation}
where $v_i$ and $\hat{v}_i$ denote the ground-truth and predicted coefficient values at frame $i$, and $N$ is the total number of frames. Lower values indicate higher coefficient accuracy.

\paragraph{Fréchet Distance (FD).} FD evaluates the similarity between the real ($P_r$) and generated ($P_g$) feature distributions under a multivariate Gaussian assumption. Given means $\mu$ and covariances $\Sigma$, it is defined as:
\begin{equation}
  \text{FD}(P_r, P_g) = \lVert \mu_r - \mu_g \rVert^2 + \text{Tr}\left(\Sigma_r + \Sigma_g - 2(\Sigma_r \Sigma_g)^{1/2}\right)
\end{equation}
A lower FD indicates higher fidelity to the ground-truth data.

\paragraph{Wasserstein Inception Distance (WInD).} WInD utilizes the 2-Wasserstein distance to provide a robust evaluation metric, particularly when the underlying data manifold exhibits complex, non-Gaussian geometries. Let $\Gamma(P_r, P_g)$ denote the set of all joint distributions $\gamma(x, y)$ with marginals $P_r$ and $P_g$:
\begin{equation}
  W_2(P_r, P_g) = \left( \inf_{\gamma \in \Gamma(P_r, P_g)} \int \lVert x - y \rVert^2 \, d\gamma(x, y) \right)^{1/2}
\end{equation}
By calculating the optimal transport plan, WInD offers a highly stable assessment gradient. Lower scores imply generated sequences that are statistically indistinguishable from the ground truth.